\documentclass[runningheads]{llncs}
\usepackage{graphicx}
\usepackage{amsfonts}
\usepackage{amsmath}
\usepackage{multirow}
\usepackage{algorithmic}
\usepackage[ruled,vlined]{algorithm2e}
\usepackage[colorlinks=true, linkcolor=blue, citecolor=blue, urlcolor=blue]{hyperref}
\usepackage[misc]{ifsym}

\begin{document}
\title{SurgicalGaussian: Deformable 3D Gaussians for High-Fidelity Surgical Scene Reconstruction}
\titlerunning{SurgicalGaussian}
%
\author{
Weixing Xie\inst{1,2} \and
Junfeng Yao\inst{1,2,3}\textsuperscript{(\Letter)} \and
Xianpeng Cao\inst{1} \and 
Qiqin Lin\inst{1} \\
Zerui Tang\inst{1,2} \and
Xiao Dong\inst{4} \textsuperscript{(\Letter)} \and
Xiaohu Guo\inst{5}
}
%
\authorrunning{W. Xie et al.}
%
\institute{
\textsuperscript{1} Center for Digital Media Computing, School of Film,  School of Informatics,\\
Xiamen University, Xiamen, China\\
\textsuperscript{2} National Institute for Data Science in Health and Medicine, Xiamen University\\
\textsuperscript{3} Key Laboratory of Digital Protection and Intelligent Processing of Intangible Cultural Heritage of Fujian and Taiwan, Ministry of Culture and Tourism\\
\textsuperscript{4} Guangdong Provincial Key Laboratory IRADS and Department of Computer Science, BNU-HKBU United International College, Zhuhai, China\\
\textsuperscript{5} Department of Computer Science, The University of Texas at Dallas, Dallas, USA}
%
\maketitle              
\begin{abstract}
Dynamic reconstruction of deformable tissues in endoscopic video is a key technology for robot-assisted surgery. Recent reconstruction methods based on neural radiance fields (NeRFs) have achieved remarkable results in the reconstruction of surgical scenes. However, based on implicit representation, NeRFs struggle to capture the intricate details of objects in the scene and cannot achieve real-time rendering. In addition, restricted single view perception and occluded instruments also propose special challenges in surgical scene reconstruction. To address these issues, we develop SurgicalGaussian, a deformable 3D Gaussian Splatting method to model dynamic surgical scenes. Our approach models the spatio-temporal features of soft tissues at each time stamp via a forward-mapping deformation MLP and regularization to constrain local 3D Gaussians to comply with consistent movement. With the depth initialization strategy and tool mask-guided training, our method can remove surgical instruments and reconstruct high-fidelity surgical scenes. Through experiments on various surgical videos, our network outperforms existing method on many aspects, including rendering quality, rendering speed and GPU usage. The project page can be found at \url{https://surgicalgaussian.github.io}.

\keywords{3D Reconstruction  \and Gaussian Splatting \and Minimally Invasive Surgery.}
\end{abstract}
\section{Introduction}
In robotic-assisted minimally invasive surgery, reconstructing the surgical scene from endoscopic videos is a critical and challenging task. The reconstruction of surgical scenes can not only help doctors operate instruments more accurately, but also is the basis for a series of downstream clinical applications, such as surgical environment simulation~\cite{chong2022virtual,montana2024saramis}, robotic surgery automation~\cite{lu2021super}, and medical teaching~\cite{schmidt2024tracking}. However, the sparse viewpoints, limited movement space, topologically changing tissues and instrument occlusion in endoscopic surgery pose key challenges to dynamic reconstruction. Despite much progress recent years, reconstruction of surgical scenes by existing methods still lose intricate details.

Previous work proposed explicit discrete representations of surgical scenes, such as point clouds~\cite{li2020super,song2017dynamic,zhou2021emdq} and surfels~\cite{long2021dssr,ma2019real}. These methods usually compensate tissue deformation by sparse warp fields, limiting their ability to handle drastic motions and color alteration due to topology changes. With the development of neural radiance fields (NeRF)~\cite{dynamiccheck,mildenhall2020nerf}, continuous representations of dynamic scene demonstrated superiority in generating high-quality appearance and geometry. Specifically, EndoNeRF~\cite{endonerf} improved dynamic NeRF framework~\cite{pumarola2021d} to reconstruct stationary monocular endoscopic scenes with depth supervision. EndoSurf~\cite{zha2023endosurf} focused on surface reconstruction by employing SDF fields~\cite{lightneus} and radiance fields to model surface dynamics and appearance. In addition, LerPlane~\cite{yang2023neural} draws on the design of feature planes~\cite{fridovich2023k} to efficiently encode space-temporal features of sampling points, significantly reducing the workload of dynamic tissue modeling. However, implicit representation of NeRFs require dense sampling on millions of rays to adapt the implicit function to the surgical scene. This consumes huge computational resources, especially in surgical scenarios with complex motion and high resolution, even with accelerated NeRF versions~\cite{li2023nerfacc,yang2023neural}. It is difficult for current work to achieve high-quality reconstruction and real-time rendering of surgical scenes at the same time.

Recently, 3D Gaussian Splatting (3DGS)~\cite{kerbl20233d} has emerged as a viable alternative 3D representation to NeRF as it yields realistic rendering while being significantly faster to train than NeRFs. Specifically, 3DGS represents the scene as anisotropic 3D Gaussians and adopts differentiable rasterization pipeline to render images. In this paper, we propose a deformable 3DGS framework tailored for endoscopic videos to reconstruct dynamic surgical scene and remove occluded instruments. To model motion fields, some methods~\cite{endogaussian,wu20234dgaussians,yang2023neural} employ planar structures for feature encoding efficiency. Although decomposing 3D scene into feature planes can speed up and improve the reconstruction quality, these low-rank planar components are not the best choice for encoding complex motion fields. Dynamic scenes possess a higher rank compared to static scenes, and explicit point-based rendering  further elevates the rank of the scene~\cite{chen2022tensorf,yang2023deformable}. In our method, given 3D Gaussians in canonical space, we utilize a deformation network to decouple the motion and geometry of surgical scene, predicting flexible Gaussian motion in observation space.

Compared with existing methods for stereo 3D reconstruction in robotic surgery, our contributions are summarized as follows: 1) We propose an deformable 3D Gaussians framework (SurgicalGaussian) for high-fidelity surgical scene reconstruction in endoscopic video; 2) we propose an efficient Gaussian initialization strategy (GIDM) to use geometry prior based on depth and mask to reduce the  motion-appearance ambiguity in single viewpoint; 3) we address the color prediction of occluded areas and the noise of Gaussian deformation fields by using color and deformation regularization respectively; 4) our method demonstrates compelling reconstruct quality with real-time rendering speed, preserving high-frequency details of tissues while removing surgical tools.

\begin{figure}[t]
\centering
 \includegraphics[width=1.0\linewidth]{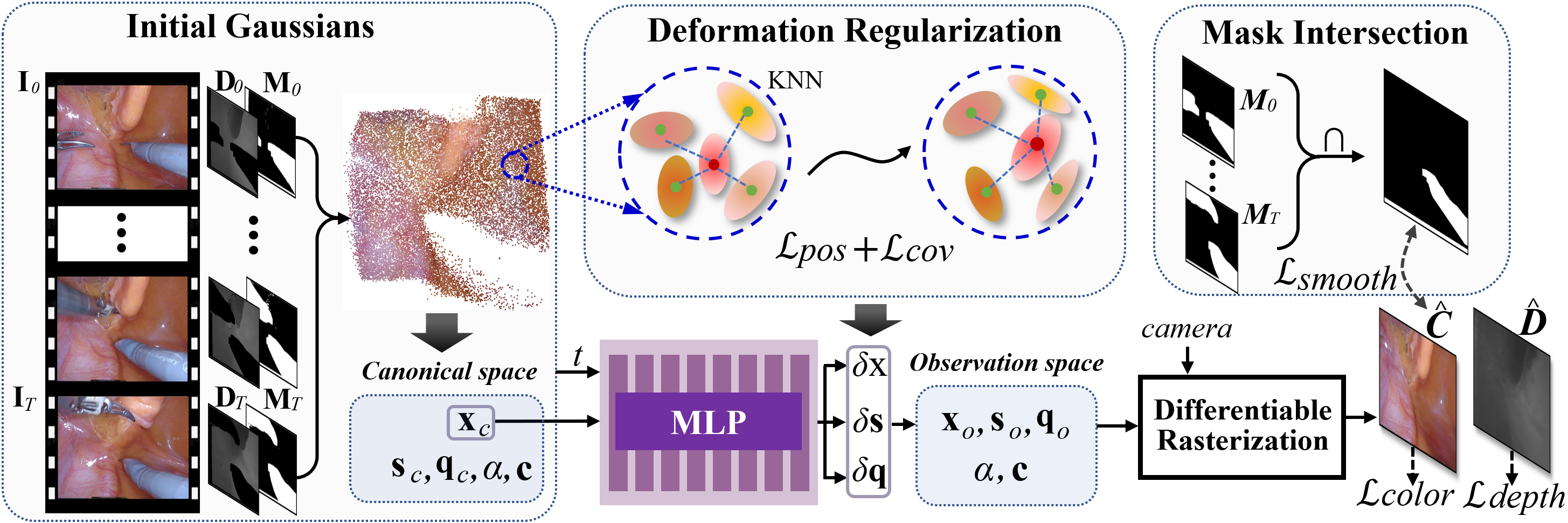}\\[-1em]
  \caption{Framework of the proposed SurgicalGaussian.}
  \label{Fig1}
\end{figure}
\section{Method}
The architecture of our network SurgicalGaussian is shown in Fig.~\ref{Fig1}. We take an endoscopic video $V=\{\mathbf{I}_i, \mathbf{D}_i, \mathbf{M}_i : i \in [0, T]\}$ as input, where $\mathbf{I}_i$ is the $i$-th frame image, $\mathbf{D}_i$ is the depth map and $\mathbf{M}_i$ (1 for tool pixels) is the mask of the surgical tools. The time $t$ of frame $i$ is normalized to $i/T \in [0, 1]$. Given the above inputs, our network builds a deformable 3D Gaussian representation of a surgical scene that can remove surgical instruments and restore deformed soft tissue with high quality.
\subsection{Preliminaries}

We build dynamic scene representation based on 3DGS~\cite{kerbl20233d}, which consists of a set of Gaussian primitives $\{\mathcal{G}\}$. The distribution of Gaussian in world space is defined by its center location $\mathbf{x}$ and the covariance matrix $\mathbf{\Sigma}$, denoted as $G(x)=exp(-1/2x^T\mathbf{\Sigma}^{-1} x)$. To ensure positive semi-definite property of Gaussians, the covariance is decomposed as $\mathbf{\Sigma} = \mathbf{RS} \mathbf{S}^T \mathbf{R}^T$, where $\mathbf{S}$ is a scaling matrix and $\mathbf{R}$ is a rotation matrix. In practice, we store the diagonal vector $\mathbf{s}\in \mathbb{R}^3 $ of the scaling matrix and the quaternion vector $\mathbf{q} \in \mathbb{R}^4$ of the rotation matrix. Each Gaussian has opacity $\alpha$ and spherical harmonics coefficients for color. We then denote a 3D Gaussian with its properties as \(\mathcal{G} = \{(\mathbf{x}, \mathbf{s}, \mathbf{q}, \alpha, \mathbf{c})\}\).

Given a viewing transformation $\mathbf{V}$ and the Jacobian
of the affine approximation of the projective transformation $\mathbf{J}$, we can project 3D Gaussians to 2D image plane for rendering. The 2D covariance matrix in camera coordinates is calculated as follows:
 $\mathbf{\Sigma}' = \mathbf{JV}\mathbf{\Sigma}\mathbf{V}^T\mathbf{J}^T$.
The estimated color $\mathit{\hat{C}}(\mathbf{r})$ and depth $\mathit{\hat{D}}(\mathbf{r})$ of pixel $\mathbf{r}$ can be rendered by blending Gaussians sorted with their depth:
\begin{equation}\label{eq1}
\begin{split}
\hat{{C}}(\mathbf{r}) = \sum_{i} (\alpha_i' \prod_{j=1}^{i-1} (1-\alpha'_j)) c_i, \hat{{D}}(\mathbf{r}) = \sum_{i} ( \alpha_i' \prod_{j=1}^{i-1} (1-\alpha'_j)) d_i. 
\end{split}
\end{equation}
Here $c_i$ is the color of 3D Gaussian, $\alpha_i'= \alpha_i\mathcal{G}_i'$ is opacity $\alpha_i$ weighted by the probability density of projected 2D Gaussian $\mathcal{G}_i'$ at target pixel location. We denote depth of $\mathcal{G}_i$ as $d_i$ and predict pixel depth in a similar way.

\subsection{GIDM initialization strategy}\label{sec.2.2}
3DGS~\cite{kerbl20233d} uses SfM~\cite{sfm} point cloud as the initial position of the 3D Gaussians. Compared with random initialization, this strategy significantly improves the rendering quality of areas not well covered by the training view. However for endoscopic videos, due to limited observation viewpoints, sparse textures of soft tissues and dynamic lighting condition, it is difficult to obtain accurate SfM point clouds, and therefore cannot provide accurate initialization for 3D Gaussians. 

In our implementation, we propose an efficient Gaussian initialization scheme using depth maps of the surgical scene. Given the depth $\mathbf{D}$ and mask $\mathbf{M}$ of T frames in surgical video, we project image pixels to world coordinates. Specifically, we first project the tissue pixels on frame 0 into 3D space to obtain the point cloud $\mathbf{P}_0$. This point cloud has large missing due to the removal of surgical tools. However, the tissue occluded by the instrument at frame $i$ may be visible in other frames, so Gaussian points should also be placed in this area for reconstruction of dynamic part. Based on this observation, we check masks in all frames and collect new tissue pixels appear in frame $i+1$ while occluded in previous $i$ frames. We add the collected pixels from other frames to frame 0 to obtain the refined image $\mathbf{I}^{\ast}$, depth map $\mathbf{D}^{\ast}$ and mask $\mathbf{M}^{\ast}$. Based on the projection, we obtain the refined point cloud $\mathbf{P}^{\ast}$. There are still holes in $\mathbf{P}^{\ast}$ if a small region is occluded the whole time.
\begin{equation}\label{eq2}
\begin{split}
\mathbf{P}^{\ast} = \{\mathbf{D}^{\ast} \mathbf{K}^{-1}_{e} \mathbf{K}^{-1}_{i} (\mathbf{I}^{\ast} \odot (\mathbf{1}-\mathbf{M}^{\ast}))\}, \mathbf{M}^* = \bigcap_{i=0}^{T} \mathbf{M}_i.
\end{split}
\end{equation}

\noindent Here, $\mathbf{K}_{i}$ and $\mathbf{K}_{e}$ are camera intrinsic matrix and extrinsic matrix respectively. We show point cloud $\mathbf{P}^{\ast}$ in Fig.~\ref{Fig1}, which is used to initialize the position \(\mathbf{x}_c\) and color \(\mathbf{c}\) of Gaussians  \(\{\mathcal{G}_c\} = \{(\mathbf{x}_c, \mathbf{s}_c, \mathbf{q}_c, \alpha, \mathbf{c})\}\) in canonical space.
\subsection{Deformable 3D Gaussian Representation}
We utilize the strong ability of 3D Gaussians on rendering to get high-fidelity reconstruction of surgical scenes. To model dynamic 3D Gaussians that vary over time, we decouple the 3D Gaussians and the deformation field. The Gaussians in canonical space represent geometric prior of the scene, and a deformation network models the changes in position and shape of the Gaussians. This Gaussian point-based representation is flexible in capturing high-rank motion of objects. 

The key to our Gaussian deformation modeling is a pure MLP. The network first encodes Gaussian position in the canonical space and time $t$ of current frame and takes them as the input of the MLP network. The deformation network $\mathbf{F}_\mathrm{\Theta}$ will learn the offset of each Gaussian's properties in the observation space, such as position \(\delta \mathbf{x}\), scaling \(\delta \mathbf{s}\) and rotation \(\delta \mathbf{q}\) to encode the motion of the scene. $\mathbf{F}_\mathrm{\Theta}$ is a deep MLP with multiple layers and $\gamma(\cdot)$ is the function for Gaussian positional encoding and time encoding using specific frequency~\cite{mildenhall2020nerf}. The offset of Gaussian position, scaling and rotation properties are obtained as following:
\begin{equation}\label{eq3}
\begin{split}
(\delta \mathbf{x}, \delta \mathbf{s}, \delta \mathbf{q}) & = \mathbf{F}_\mathrm{\Theta}((\gamma(\mathbf{x}_c), \gamma(t)).
\end{split}
\end{equation}
Then the properties of Gaussians \(\{\mathcal{G}_o\} = \{(\mathbf{x}_o, \mathbf{s}_o, \mathbf{q}_o, \alpha, \mathbf{c})\}\) in observation space are obtained:
\begin{equation}\label{eq4}
\begin{split}
\mathbf{x}_o = \mathbf{x}_c + \delta \mathbf{x}, \mathbf{s}_o = \mathbf{s}_c \cdot \exp(\delta \mathbf{s}), \mathbf{q}_o = \mathbf{q}_c \cdot \delta \mathbf{q}.
\end{split}
\end{equation}

\noindent Note that applying the $\cdot$ operation to quaternion vectors is equivalent to multiplying the corresponding rotation matrices. The deformation network $\mathbf{F}_\mathrm{\Theta}$ does not encode $\alpha$ and $\mathbf{c}$ because they are intrinsic properties of Gaussians that do not change with motion.

\subsection{Optimization}
Under the supervision of reconstruction losses and regularization terms, our network jointly optimizes canonical Gaussians $\mathcal{G}_c$ and parameters of deformation network. Since the surgical instruments in the video are to be removed, we invert the masks to select soft tissues and apply the reconstruction loss to this area. $\hat{\mathbf{C}}_i$ and $\hat{\mathbf{D}}_i$ are rendered image and depth for $i$-th frame, respectively.

\begin{equation}\label{eq5}
\begin{split}
\mathcal{L}_{\text{color}} = \left\| (\mathbf{I}_i - \hat{\mathbf{C}}_i)(\mathbf{1}-\mathbf{M}_i) \right\|_1, \mathcal{L}_{\text{depth}} = \left\| (\mathbf{D}_i - \hat{\mathbf{D}}_i)(\mathbf{1}-\mathbf{M}_i) \right\|_1.
\end{split}
\end{equation}

\textbf{Deformation regularization.} When dealing with a single-view input, the deformation network can be highly under-constrained, leading to noisy deformations from the canonical space to the observation space. To address this issue, we propose a regularization method that ensures nearby Gaussians have similar deformation~\cite{gstracking,tagliabue2021intra}. Specifically, we employ deformation consistency loss to the Gaussian position $\mathbf{x}$ and covariance matrix $\mathbf{\Sigma}$. 
For a Gaussian $\mathcal{G}_i$, we collect its $K$ ($K$ is set to 5) nearest neighbor Gaussians, calculate Euclidean distance between $\mathcal{G}_i$ and its neighbors in both canonical space and observation space, and constrain the difference after deformation to ensure consistent movement.
\begin{equation}\label{eq6}
\begin{split}
\mathcal{L}_{pos} = \sum_{i=1}^{N} \sum_{k=1}^{K}\left\|d\left(\mathbf{x}_{c}^{(i)}, \mathbf{x}_{c}^{(k)}\right)-d\left(\mathbf{x}_{o}^{(i)}, \mathbf{x}_{o}^{(k)}\right)\right\|_1,
\end{split}
\end{equation}
\begin{equation}\label{eq7}
\begin{split}
\mathcal{L}_{cov} = \sum_{i=1}^{N} \sum_{k=1}^{K}\left\|d\left(\mathbf{\Sigma}_{c}^{(i)}, \mathbf{\Sigma}_{c}^{(k)}\right)-d\left(\mathbf{\Sigma}_{o}^{(i)}, \mathbf{\Sigma}_{o}^{(k)}\right)\right\|_1.
\end{split}
\end{equation}

\textbf{Occlusion-based color regularization.} For the occlusion of surgical instruments, scene representation based on NeRFs can naturally complement the color of the occluded area benefit from the smoothness of MLP~\cite{endonerf}.  The representation based on 3DGS is discrete and cannot handle rendering on occluded regions well. We observed that occlusions caused by surgical instruments fall into two types, one visible in other frames and the other invisible in all frames. The mask $\mathbf{M}^{\ast}$ discussed in Sec.~\ref{sec.2.2} represents the occluded area that has never been observed. Notice that a Gaussian located in this region will hardly be optimized at any time, thus generating holes in rendered images. In order to enable the Gaussians in occluded regions to learn color similar to those of nearby Gaussians, we introduce a total variational loss~\cite{fridovich2023k}, which helps generate reasonable color in occluded regions. The color regularization loss is defined as:
\begin{equation}\label{eq8}
\begin{split}
\mathcal{L}_{smooth} = \frac{1}{n} \sum_{p,q} 
    \big(\lVert \textbf{C}^{p,q} - \textbf{C}^{p-1,q}\rVert_2^2 +
        \lVert \textbf{C}^{p,q} - \textbf{C}^{p,q-1}\rVert_2^2 \big),
\mathbf{C}=\hat{\mathbf{C}}_i \odot \mathbf{M}^*,
\end{split}
\end{equation}
where $p$ and $q$ are indices of pixel, and $n$ denotes the number of pixels in $\mathbf{C}$.
The selection of intersection mask $\mathbf{M}^{\ast}$ instead of mask of each frame encourages Gaussians to use the ground truth appearance to recover temporarily occluded areas, thus preventing tissue color in that area from being over smoothed.

\textbf{Total loss.} We combine reconstruction loss and regularization loss to optimize dynamic 3D Gaussian representation. Follow 3DGS~\cite{kerbl20233d}, we add SSIM loss to ensure structural similarity of rendered image to ground-truth image. We use five parameters to balance the relative importance of different terms. The final optimization objective can be represented as follows: 
\begin{equation}\label{eq9}
\begin{split}
\mathcal{L} = (\mathcal{L}_{color}+ \lambda_{\text{1}}\mathcal{L}_{ssim}+\lambda_{\text{2}}\mathcal{L}_{depth})+(\lambda_{\text{3}}\mathcal{L}_{pos}+\lambda_{\text{4}}\mathcal{L}_{cov}+\lambda_{\text{5}}\mathcal{L}_{smooth}).
\end{split}
\end{equation}

\section{Experiments}
\subsection{Experimental Settings}

\begin{table}[!t]
\caption{Quantitative comparisons of our method with EndoNeRF~\cite{endonerf}, EndoSurf~\cite{zha2023endosurf}, LerPlane~\cite{yang2023neural} and EndoGaussian~\cite{endogaussian}. We evaluate reconstruction quality on PSNR↑, SSIM↑, LPIPS↓ together with rendering speed FPS↑ and GPU Usage↓.}

\centering
\begin{tabular}{c|c|ccc|ccc|cc}
\noalign{\smallskip}
\hline
\multirow{2}{*}{Dataset}
& \multirow{2}{*}{Methods}
& \multicolumn{3}{c|}{\textquotedblleft pulling\textquotedblright} & \multicolumn{3}{c|}{\textquotedblleft cutting\textquotedblright} & \multirow{2}{*}{FPS} & \multirow{2}{*}{GPU}\\ \cline{3-8}
& & PSNR & SSIM & LPIPS  & PSNR & SSIM & LPIPS \\ \hline
\multirow{5}{*}{EndoNeRF} & EndoNeRF & 34.217 & 0.938 & 0.160 & 34.186 & 0.932 & 0.151  & 0.04 & 15 GB\\
 & EndoSurf & 35.004 & 0.956 & 0.120 & 34.981 & 0.953 & 0.106 & 0.05 & 17 GB\\
 & LerPlane & 36.241 & 0.950 & 0.102 & 35.580 & 0.955 & 0.101 & 1.02 & 20 GB\\
 & EndoGaussian & 37.308 & 0.958 & 0.070 & \textbf{38.287} & \textbf{0.962} & \textbf{0.058} & \textbf{190} & \textbf{2 GB}\\
 & Ours & \textbf{38.783} & \textbf{0.970} & \textbf{0.049} & 37.505 &  0.961 & 0.062 & 80 & 4 GB \\ \hline
 \multirow{6}{*}{StereoMIS} &  & \multicolumn{3}{c|}{\textquotedblleft intestine\textquotedblright} & \multicolumn{3}{c|}{\textquotedblleft liver\textquotedblright}\\ \cline{2-10}
 & EndoNeRF & 28.694 & 0.783 & 0.279 & 27.738 & 0.712 & 0.345  & 0.06 & 13 GB\\
 & EndoSurf & 29.660 & 0.853 & 0.204 & 28.941 & 0.820 & 0.248 & 0.08 & 14 GB\\
 & LerPlane & 29.441 & 0.822 & 0.206 & 28.852 & 0.793 & 0.254 & 1.45 & 19 GB\\
 & EndoGaussian & 29.024 & 0.805 & 0.213 & 26.174 & 0.728 & 0.295 & \textbf{200} & \textbf{2 GB}\\
 & Ours & \textbf{31.496} & \textbf{0.890} & \textbf{0.145} & \textbf{31.668} &  \textbf{0.893} & \textbf{0.135} & 140 & 3 GB \\
\hline
\end{tabular}
\label{tab:experiment}
\end{table}

\begin{table}[!t]
\caption{Ablation study on EndoNeRF~\cite{endonerf} dataset.}
\centering

\begin{tabular}{@{\hspace{0.5cm}}p{3.5cm}|ccc|ccc}
\noalign{\smallskip}
\hline
\multirow{2}{*}{Model}
& \multicolumn{3}{c|}{\textquotedblleft pulling\textquotedblright} & \multicolumn{3}{c}{\textquotedblleft cutting\textquotedblright} \\ \cline{2-7}
& PSNR & SSIM & LPIPS  & PSNR & SSIM & LPIPS \\ \hline
w/o GIDM initialization & 37.771 & 0.962 & 0.091 & 36.428 & 0.944 & 0.089 \\
w/o $\mathcal{L}_{pos}$,$\mathcal{L}_{cov}$ & 37.944 & 0.963 & 0.094 & 36.633 & 0.951 & 0.090\\
w/o $\mathcal{L}_{smooth}$ & 38.609 & 0.968 & 0.070 & 37.081 & 0.957 & 0.065\\
Full model & \textbf{38.783} & \textbf{0.970} & \textbf{0.049} & \textbf{37.505} &  \textbf{0.961} & \textbf{0.062}\\ \hline
\end{tabular}
\label{tab:ablation}
\end{table}

\textbf{Datasets and evaluation metrics.} The EndoNeRF dataset\cite{endonerf} contains two prostatectomy cases shot by a binocular camera, with endoscopic images, depth maps, and mask maps of surgical tools. 
We select two sequences with rapid tissue motion from StereoMIS \cite{StereoMIS} dataset to evaluate reconstruction ability of existing methods. 
We estimate depth maps using a pre-trained STTR-light \cite{sttr} model, and generate tool masks based on SAM-Track \cite{samtrack} model.
We adopt three metrics to measure the quality of surgical scene reconstruction, including PSNR, SSIM, and LPIPS. These metrics are calculated by comparing predicted color and ground-truth color excluding surgical instruments denoted by masks. We also report GPU memory usage for training and rendering speed FPS.

\begin{figure}[t]
\centering
\begin{minipage}[b]{1\linewidth}
\hspace{1.6em}
\fontsize{7.0pt}{\baselineskip}\selectfont EndoNeRF
\hspace{1.5em}
\fontsize{7.0pt}{\baselineskip}\selectfont EndoSurf
\hspace{2.2em}
\fontsize{7.0pt}{\baselineskip}\selectfont LerPlane
\hspace{1.2em}
\fontsize{7.0pt}{\baselineskip}\selectfont EndoGaussian
\hspace{1.8em}
\fontsize{7.0pt}{\baselineskip}\selectfont Ours
\hspace{3.2em}
\fontsize{7.0pt}{\baselineskip}\selectfont Reference
\end{minipage}
\begin{minipage}[b]{1.0\linewidth}
\centering
\includegraphics[width=0.95\linewidth]{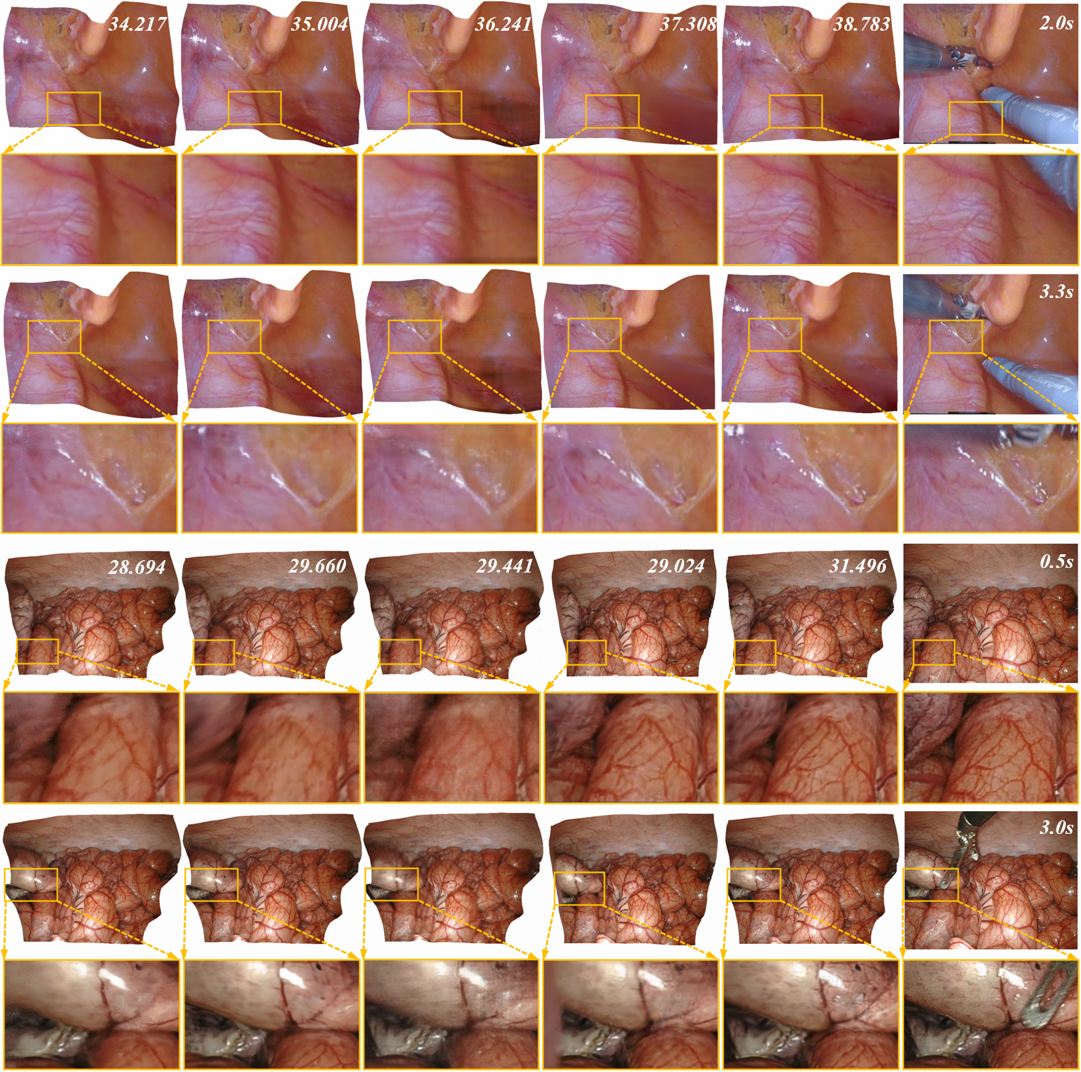}\\[-1em]
\end{minipage}\\[0.2em]
\caption{Comparison of reconstruction results between our SurgicalGaussian and EndoNeRF~\cite{endonerf}, EndoSurf~\cite{zha2023endosurf}, LerPlane~\cite{yang2023neural} and EndoGaussian~\cite{endogaussian}.}
\label{Fig2}
\end{figure}

\textbf{Implementation details.} The deformation MLP $\mathbf{F}_\mathrm{\Theta}$ has 8 hidden layers and a width of 256. We randomly select one frame for each training iteration, and train all scenes with $40k$ iterations. The loss weights in Eq.\eqref{eq9} are empirically set as \(\lambda_1 = 0.2\), \(\lambda_2 = 0.001\), \(\lambda_3 = 1\), \(\lambda_4 = 200\) and \(\lambda_5 = 0.02\). Adam~\cite{adam} optimizer is adopted for training, the initial learning rate of the deformation network is set to \(1.5 \times 10^{-5}\), and the Gaussian optimization parameter settings follow 3DGS~\cite{kerbl20233d}. We conduct all experiments on a computer with an RTX 3090 graphics card.

\subsection{Comparisons \& Ablations}

\textbf{Comparison experiments.} We conduct comparison experiments of SurgicalGaussian and other reconstruction methods on two datasets. Dynamic NeRFs, such as EndoNeRF~\cite{endonerf}, LerPlane~\cite{yang2023neural} and EndoSurf~\cite{zha2023endosurf}, are based on an inverse mapping from observation space to canonical space when model the motion field, which is unable to achieve high-quality decoupling of canonical space and deformation field~\cite{yang2023deformable}. In addition, EndoNeRF uses MLP to encode both motion and appearance of the scene, fails to capture high-frequency details. LerPlane fixes the position of the sampling points on the grid nodes for encoding, and the resulting neural features cannot adapt well to high-rank complex signals. EndoSurf employs smoothness constraint on reconstructed dynamic surfaces, thus compromising rendering quality. We also compare our network with a similar 3DGS-based reconstruction method EndoGaussian~\cite{endogaussian}, which strives for efficient training speed and real-time rendering. However, EndoGaussian faces the same problem as LerPlane, using feature plane, the low-rank tensors, to encode Gaussian deformation fields, resulting in impaired reconstruction of objects moving rapidly. In Table~\ref{tab:experiment} and Fig.~\ref{Fig2}, we give quantitative evaluations and reconstructed point clouds of these methods. Our method can capture the intricate details of objects and produce highly realistic rendering results.

\textbf{Rendering efficiency.} Our method achieves real-time rendering. Although the introduction of deformation MLP may increase the rendering overhead, we are still able to render in real-time owing to the extremely efficient CUDA implementation of 3D Gaussian splatting and our compact MLP structure. The average rendering speed on RTX3090 is $\geq$ 80FPS and GPU usage is $\leq$ 4GB.

\textbf{Ablations.} In Table~\ref{tab:ablation}, we conduct ablation analysis on the effectiveness of different modules of our network. We compare reconstruction quality based on random and GIDM initialization. GIDM helps restore the scene geometry thus yielding better quality. The $\mathcal{L}_{pos}$ and $\mathcal{L}_{cov}$ losses smooth tissue surface deformations by reducing overly noisy Gaussian motion fields. Although $\mathcal{L}_{smooth}$ loss slightly improves the quality of the scene, it plays an important role in restoring the color of occluded areas by surgical tools.

\section{Conclusion}

We present a novel deformable 3D Gaussian Splatting framework, SurgicalGaussian, specifically designed for high-fidelity reconstruction of dynamic surgical scenes. The 3D Gaussian-based representation in canonical space captures intricate textures of the tissue, while the forward-mapping deformation field enhances its ability to model complex motions. The ablation study demonstrates the effectiveness of the proposed modules in our network, such as depth initialization, deformation regularization and color smoothness. We conduct extensive experiments, and our method significantly outperforms the SOTA methods in reconstruction quality. Our method provides a new idea for the reconstruction of surgical scenes and will further promote the development of robot-assisted surgery and intelligent medical care.
\newpage

%
%
%
\bibliographystyle{splncs04}
\bibliography{reference}
%
\end{document}